\documentclass[sigconf]{acmart}

\renewcommand\footnotetextcopyrightpermission[1]{}
\setcopyright{none}

\AtBeginDocument{%
  }

\usepackage{microtype}
\usepackage{cleveref}
\usepackage{algorithm}
\usepackage[noend]{algpseudocode}
\usepackage{mathtools}

\newcommand{\searchSpace}{\mathcal{X}}
\newcommand{\individual}{\mathbf{x}}
\newcommand{\population}{\mathbf{X}}

\newcommand{\fitness}{f}
\newcommand{\fitnesses}{\mathbf{\fitness}}

\newcommand{\descriptorSpace}{\mathcal{D}}
\newcommand{\descriptor}{\mathbf{d}}
\newcommand{\descriptors}{\descriptor}

\newcommand{\centroid}{\mathbf{c}}

\newcommand{\set}[2]{\left\{#1 \, \middle| \, #2\right\}}
\newcommand{\norm}[1]{\left\lVert#1\right\rVert}

\definecolor{red}{RGB}{175, 42, 20}
\definecolor{green}{RGB}{109, 158, 75}
\definecolor{blue}{RGB}{76, 115, 187}
\definecolor{purple}{RGB}{90, 64, 154}

\begin{document}

\title[Discovering Quality-Diversity Algorithms via Meta-Black-Box Optimization]{Discovering Quality-Diversity Algorithms via\texorpdfstring{\linebreak}{} Meta-Black-Box Optimization}

\author{Maxence Faldor}
\email{m.faldor22@imperial.ac.uk}
\orcid{0000-0003-4743-9494}
\affiliation{%
    \institution{Imperial College London}
    \city{London}
    \country{United Kingdom}
}

\author{Robert Tjarko Lange}
\email{robert@sakana.ai}
\orcid{0009-0005-8799-1138}
\affiliation{%
    \institution{Sakana AI}
    \city{Tokyo}
    \country{Japan}
}

\author{Antoine Cully}
\email{a.cully@imperial.ac.uk}
\orcid{0000-0002-3190-7073}
\affiliation{%
    \institution{Imperial College London}
    \city{London}
    \country{United Kingdom}
}

\renewcommand{\shortauthors}{Faldor et al.}

\begin{abstract}
Quality-Diversity has emerged as a powerful family of evolutionary algorithms that generate diverse populations of high-performing solutions by implementing local competition principles inspired by biological evolution.
While these algorithms successfully foster diversity and innovation, their specific mechanisms rely on heuristics, such as grid-based competition in MAP-Elites or nearest-neighbor competition in unstructured archives.
In this work, we propose a fundamentally different approach: using meta-learning to automatically discover novel Quality-Diversity algorithms. By parameterizing the competition rules using attention-based neural architectures, we evolve new algorithms that capture complex relationships between individuals in the descriptor space.
Our discovered algorithms demonstrate competitive or superior performance compared to established Quality-Diversity baselines while exhibiting strong generalization to higher dimensions, larger populations, and out-of-distribution domains like robot control.
Notably, even when optimized solely for fitness, these algorithms naturally maintain diverse populations, suggesting meta-learning rediscovers that diversity is fundamental to effective optimization.
\end{abstract}


\begin{teaserfigure}
    \includegraphics[width=\textwidth]{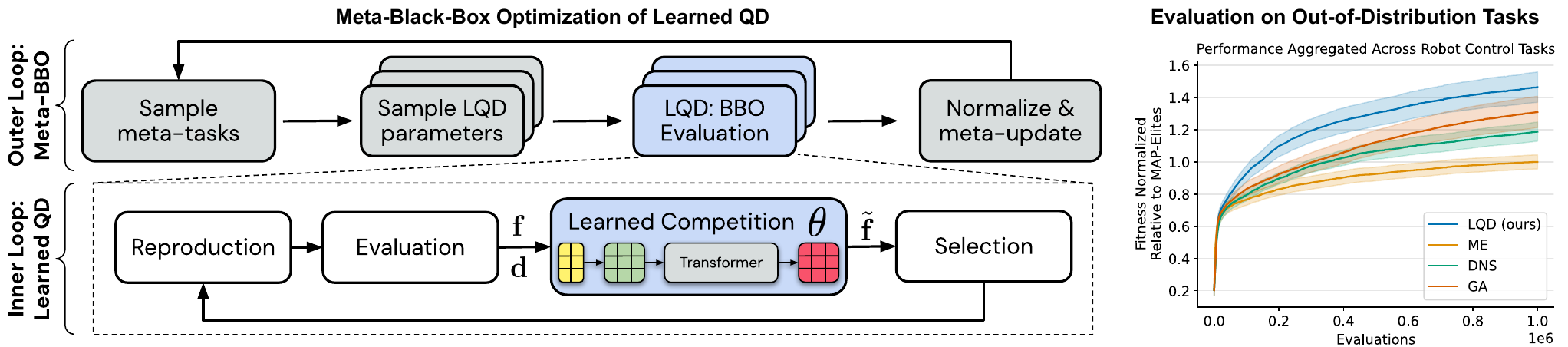}
    \caption{Meta-black-box optimization of Learned Quality-Diversity (LQD) algorithms. At each meta-generation, a meta-evolution strategy (ES) samples candidate LQD parameters and evaluates them on sampled black-box optimization tasks. The LQD algorithm is run, performance is normalized across tasks, and the meta-ES is updated accordingly. After meta-optimization the discovered LQD outperforms various baselines across 6 aggregated robot control tasks.}
    \Description{A schematic diagram showing the nested optimization process. The outer loop depicts the meta-evolution strategy sampling and updating algorithm parameters. The inner loop shows the evaluation of each candidate algorithm on multiple optimization tasks, with performance metrics flowing back to inform the outer loop updates.}
    \label{fig:teaser}
\end{teaserfigure}

\received{29 January 2025}
\received[revised]{12 March 2009}
\received[accepted]{5 June 2009}

\maketitle

\pagestyle{plain}

\section{Introduction}
Over billions of years, biological evolution has created the extraordinary complexity of life on Earth, giving rise to the flight of birds, the computational power of the human brain, and the collaborative intelligence of ant colonies.
This remarkable process inspired the development of Genetic Algorithms~\citep{ga} (GA), which replicate the essential principles of biological evolution in silico to solve challenging computational problems.
In their classic implementation, GAs maintain a population of candidate solutions that evolve over generations through reproduction, evaluation and selection.
However, this process typically implements selection through global competition, where each individual competes against the entire population for survival. Such a population-wide selection pressure often leads to premature convergence toward a single suboptimal solution, losing the diversity that makes natural evolution so powerful~\citep{mouret_IlluminatingSearchSpaces_2015}.

In contrast, natural evolution is driven by local competition, where individuals compete primarily with neighbors sharing the same environment~\citep{darwin}.
This localized competitive pressure naturally leads to the emergence of diverse specialized organisms. When success is defined by outperforming immediate neighbors rather than the entire population, species can adapt to their specific niches without being eliminated by globally superior solutions. Each local environment presents unique challenges and opportunities, fostering specialized traits optimal for that context, rather than forcing convergence to a single dominant species.

The power of local competition in driving both diversity and innovation has been harnessed in Quality-Diversity algorithms~\citep{pugh_QualityDiversityNew_2016} (QD), a family of methods that evolve populations of diverse, high-performing solutions.
Notably, maintaining diversity has been shown to improve not just population-level adaptation but also peak performance: by promoting exploration and serendipitous discoveries, the best solutions emerge from diverse populations rather than pure fitness optimization~\citep{greatness,faldor_MAPElitesDescriptorConditionedGradients_2023,faldor_SynergizingQualityDiversityDescriptorConditioned_2024}. This counter-intuitive benefit arises because preserving diverse solutions, even when initially suboptimal, creates crucial stepping stones for evolution. These intermediate solutions can later evolve into high-performing variants that would have been inaccessible through greedy optimization.

To implement local competition, QD algorithms typically embed individuals in a vector space --- called the descriptor space --- where distances between vectors determine which solutions compete with each other and what constitutes a novel solution. MAP-Elites~\citep{mouret_IlluminatingSearchSpaces_2015} (ME), a prominent QD algorithm, implements local competition by discretizing this descriptor space into a grid of cells, where competition occurs only between solutions mapped to the same cell.
Alternative approaches maintain unstructured archives of solutions where competition occurs between similar individuals without imposing a rigid spatial organization~\citep{lehman_AbandoningObjectivesEvolution_2011,faldor_ArtificialOpenEndedEvolution_2024,cully_AutonomousSkillDiscovery_2019,dns}.
While these Quality-Diversity algorithms successfully implement local competition, their particular mechanisms often rely on heuristics. For instance, ME's grid-based competition rely on a rather crude discretization of the descriptor space~\citep{dns,cqd,vassiliades_ComparisonIlluminationAlgorithms_2017}, while unstructured archives often rely on additional hyperparameters that prove particularly challenging to tune~\citep{csc}.

In this work, we explore a fundamentally different approach: given a sufficiently flexible parameterization of the competition rules, we discover new Quality-Diversity algorithms using meta-black-box optimization~\citep{lange_DiscoveringAttentionBasedGenetic_2023,lange_DiscoveringEvolutionStrategies_2023,lange_EvolutionTransformerInContext_2024}.
Our key contributions include: (1) Learned Quality-Diversity (LQD), a flexible framework that parameterizes competition rules using attention-based neural architectures, enabling the discovery of sophisticated Quality-Diversity algorithms; (2) empirical evidence that our learned algorithms match or exceed the performance of established baselines like MAP-Elites~\citep{mouret_IlluminatingSearchSpaces_2015} and Dominated Novelty Search~\citep{dns}; and (3) evidence of strong generalization capabilities, with LQD scaling effectively to higher dimensions, larger populations, and novel domains like robot control tasks. Our results suggest that meta-learning can automatically discover sophisticated local competition strategies that advance the state-of-the-art in Quality-Diversity optimization.

\vspace{-0.065cm}
\section{Related Work}
\textbf{Diversity in Genetic Algorithms.}
Evolution's remarkable capacity for generating diverse, well-adapted species has inspired numerous algorithmic approaches in evolutionary computation. Early diversity maintenance techniques focused on preventing premature convergence through mechanisms like fitness sharing~\citep{GeneticAlgorithmsSharing1987} and hierarchical fair competition~\citep{hfc}, which restrict competition to occur within species or niches. The NEAT algorithm~\citep{neat} demonstrated the power of this approach by using speciation to protect structural innovations in evolving neural networks, allowing novel architectures to optimize before competing with established solutions. Novelty Search~\citep{lehman_AbandoningObjectivesEvolution_2011} took a radical step by completely abandoning the objective function, instead rewarding solutions solely for being different from their predecessors. This approach proved effective at avoiding deception by discovering valuable stepping stones that initially appeared suboptimal.

\textbf{Quality-Diversity.}
Building on these insights, QD optimization emerged as a distinct family of algorithms that combine novelty with local competition. Novelty Search with Local Competition~\citep{nslc} pioneered this approach by having solutions compete only against behavioral neighbors, while MAP-Elites~\citep{mouret_IlluminatingSearchSpaces_2015} formalized it through a grid-based architecture where each cell maintains its highest-performing solution. These approaches have proven particularly valuable in robotics~\citep{cully_RobotsThatCan_2015}, where diverse behavioral repertoires enable adaptation. However, current QD algorithms often rely on simple heuristics --- grid-based approaches impose rigid discretization, while nearest-neighbor methods may miss important relationships between distant solutions. This suggests an opportunity to discover more sophisticated competition mechanisms that better capture the dynamics driving natural evolution's success.

\textbf{Discovering Algorithms via Meta-Optimization.}
Over the course of machine learning’s evolution, handcrafted components have gradually been replaced by modules learned directly from data.
Recent work continues this trend by pursuing end-to-end discovery of gradient descent-based optimizers \citep{bengio_1992, andrychowicz_2016, metz2019understanding}, objective functions in Reinforcement Learning~\citep{oh2020discovering,xu2020meta,lu2022discovered}, the online tuning of hyperparameter schedules~\citep{xu2018meta,zahavy2020self,parker2022automated}, and the meta-learning of complete learning algorithms~\citep{wang2016learning,kirsch2021meta,kirsch2022introducing}.
In these methods, appropriate inductive biases supplied by neural network parametrizations can guide and constrain the meta-search process. Here, our proposed LQD architecture leverages the equivariance properties encoded by the self-attention mechanism~\citep{lee2019set,kossen2021self,tang_SensoryNeuronTransformer_2021}, thereby defining a family of QD algorithms parameterized by neural networks.

\textbf{Population-Based Optimization via Meta-Optimization.}
Several attempts have been made to automatically optimize evolutionary algorithms. E.g. \citet{chen_2017,tv2019meta, gomes2021meta} explored meta-learning entire algorithms for low-dimensional BBO using a sequence model to process solution candidates and/or their fitness values. \citet{shala2020learning}, on the other hand, introduced a meta-optimized policy to control the scalar search scale of CMA-ES~\citep{hansen2001completely}. However, these methods often have limited generalization capabilities and are restricted to certain optimization domains, fixed population sizes, or fixed search dimensions.
Most closely related to our work are LES~\citep{lange_DiscoveringEvolutionStrategies_2023} and LGA~\citep{lange_DiscoveringAttentionBasedGenetic_2023}. Both approaches relied on the invariance of dot-product self-attention~\citep{lee2019set,tang_SensoryNeuronTransformer_2021} to generalize to unseen meta-training settings.
Here, we successfully meta-optimize a parametrized Transformer model to discover various QD algorithms, which can outperform baseline algorithms on unseen optimization tasks. 

\section{Background}
\subsection{Black-Box Optimization}
\label{sec:background-bbo}
Black-box optimization (BBO) addresses the fundamental challenge of finding optimal solutions without access to gradients or internal function structure. In this work, we focus on real-parameter black-box optimization. Given an objective function $f: \mathbb{R}^n \rightarrow \mathbb{R}$ with unknown functional form, we seek to solve:
$\max_{\individual \in \searchSpace} f(\individual)$
where the search space $\searchSpace \subset \mathbb{R}^n$ is a $n$-dimensional box.

\subsection{Genetic Algorithms}
\label{sec:background-ga}
Genetic Algorithms~\citep{ga,rechenberg_EvolutionsstrategieOptimierungTechnischer_1973} (GAs) are population-based black-box optimization methods that evolve a population of candidate solutions across generations.
These algorithms have become fundamental tools in optimization due to their ability to handle non-convex, multi-modal landscapes where traditional gradient-based methods often fail.
\Cref{alg:ga} outlines a standard GA implementation, though numerous variations exist.

\begin{algorithm}
\caption{Genetic Algorithm}
\label{alg:ga}
\small
\begin{algorithmic}
\Require population size $N$, reproduction batch size $B$
\State Initialize population $\population$ with fitness $\fitnesses$
\For{each generation}
    \State $\population^\prime \gets \textsc{reproduction}(\population, \fitnesses)$\Comment{Generate $B$ offspring}
    \State $\mathrlap{\population}\hphantom{\population^\prime} \gets \textsc{concat}(\population, \population^\prime)$\Comment{Add offspring to population}
    \State $\mathrlap{\fitnesses}\hphantom{\population^\prime} \gets \textsc{evaluation}(X)$\Comment{Evaluate fitness}
    \State {\color{red}$\mathrlap{\tilde{\fitnesses}}\hphantom{\population^\prime} \gets \textsc{competition}(\fitnesses)$\Comment{Global competition (identity)}}
    \State $\mathrlap{\population}\hphantom{\population^\prime} \gets \textsc{selection}(\population, \tilde{\fitnesses})$\Comment{Keep top-$N$ individuals}
\EndFor
\end{algorithmic}
\end{algorithm}

At each generation, a GA updates a population of $N$ solutions $\population = (\individual_i)_{i=1}^N \in \mathbb{R}^{N \times n}$ with corresponding fitness values $\fitnesses = (\fitness_i)_{i=1}^N \in \mathbb{R}^N$.
First, $B$ new offspring solutions $\population^\prime = \textsc{reproduction}(\population, \fitnesses)$ are generated through mechanisms like crossover and mutation.
These offspring are added to the population, creating an enlarged set $\textsc{concat}(\population, \population^\prime)$ of size $N+B$.
Then, $\textsc{competition}$ computes the \emph{competition fitness} $\tilde{\fitnesses}$ that will determine survival. In classic GAs, this function is simply the identity function $\tilde{\fitnesses} = \fitnesses$, meaning individuals engage in global competition based on their raw fitness values --- this baseline will be modified in QD algorithms to implement local competition.
Finally, the selection operator ranks all solutions according to their competition fitness $\tilde{\fitnesses}$ and retains only the top-$N$ solutions through truncation, forming the next generation.

\subsection{Quality-Diversity}
\label{sec:background-qd}
Quality-Diversity~\citep{pugh_QualityDiversityNew_2016} is a family of genetic algorithms that implement local competition to generate diverse, high-performing solutions.
In a QD algorithm, each solution is characterized by both its fitness value $\fitness$ and a descriptor vector $\descriptor \in \descriptorSpace \subset \mathbb{R}^D$ that captures meaningful features of the individual.
The descriptors define precisely how solutions differ from each other, providing a mathematical foundation for local competition where solutions compete primarily with similar neighbors rather than the entire population.

\begin{algorithm}
\caption{Quality-Diversity}
\label{alg:qd}
\small
\begin{algorithmic}
\Require population size $N$, reproduction batch size $B$
\State Initialize population $\population$ with fitness $\fitnesses$ and descriptors $\descriptors$
\For{each generation}
    \State $\mathrlap{\population^\prime}\hphantom{\fitnesses, \descriptors} \gets \textsc{reproduction}(\population, \fitnesses)$\Comment{Generate $B$ offspring}
    \State $\mathrlap{\population}\hphantom{\fitnesses, \descriptors} \gets \textsc{concat}(\population, \population^\prime)$\Comment{Add offspring to population}
    \State $\mathrlap{\fitnesses, \descriptors}\hphantom{\fitnesses, \descriptors} \gets \textsc{evaluation}(X)$\Comment{Evaluate fitness and descriptor}
    \State {\color{green}$\mathrlap{\tilde{\fitnesses}}\hphantom{\fitnesses, \descriptors} \gets \textsc{competition}(\fitnesses, \descriptors)$\Comment{Local competition}}
    \State $\mathrlap{\population}\hphantom{\fitnesses, \descriptors} \gets \textsc{selection}(\population, \tilde{\fitnesses})$\Comment{Keep top-$N$ individuals}
\EndFor
\end{algorithmic}
\end{algorithm}

At each generation, a QD algorithm updates a population of $N$ solutions $\population = (\individual_i)_{i=1}^N \in \mathbb{R}^{N \times n}$ with corresponding fitness values $\fitnesses = (\fitness_i)_{i=1}^N \in \mathbb{R}^N$ and descriptors $\descriptors = (\descriptor_i)_{i=1}^N$.
The key innovation lies in replacing global competition with local competition. While GAs used the identity function for $\textsc{competition}$, QD algorithms compute competition fitness values $\tilde{\fitnesses}$ based on both $\fitnesses$ and $\descriptors$. Individuals primarily compete with others having similar descriptors, creating local competitive pressures that promote both quality and diversity.
The selection of individuals for the next generation remains the same --- ranking by $\tilde{\fitnesses}$ followed by truncation --- but now operates on locally-modified fitness values. This local competition allows the population to maintain diverse solutions adapted to different niches while driving improvement within each niche.

\subsubsection{MAP-Elites}
\label{sec:background-me}
ME is a prominent QD algorithm that implements local competition by discretizing the descriptor space into a grid of cells~\citep{mouret_IlluminatingSearchSpaces_2015,vassiliades_UsingCentroidalVoronoi_2018}. Each cell in the grid is represented by a centroid $\centroid_i$ and maintains at most one solution. Competition occurs only between solutions mapped to the same cell. Given a population with fitness values $\fitnesses$ and descriptors $\descriptors$, the $\textsc{competition}$ function operates as follows:
\begin{enumerate}
    \item Each individual is assigned to its nearest centroid based on its descriptor, partitioning the population into grid cells.
    \item Within each cell $i$, only the highest-fitness individual $k$ maintains its original fitness value $\tilde{\fitness}_k = \fitness_k$, while all other individuals are eliminated by setting $\tilde{\fitness}_j = -\infty$, for $j \neq k$.
\end{enumerate}
This mechanism ensures that only the best-performing solution survives in each cell of the descriptor space, effectively implementing local competition.

\subsubsection{Novelty Search}
\label{sec:background-ns}
While not strictly a Quality-Diversity algorithm, Novelty Search (NS) pioneered key ideas that influenced QD approaches. This algorithm rewards solutions for being different an archive of past solutions, introducing a powerful mechanism for maintaining diversity. Here, we present a simplified variant where novelty is computed using only the current population. Given a population with fitness values $\fitnesses$ and descriptors $\descriptors$, the $\textsc{competition}$ function operates as follows:
\begin{enumerate}
    \item For each solution $i$, compute the descriptor space distances to all other solutions $\mathcal{D}_i = \set{\norm{\descriptor_i - \descriptor_j}}{j = 1, \dots, N}$.
    \item Compute $\tilde{\fitness}_i$ as the \textit{novelty score}: the average distance to the $k$-nearest neighbors.
\end{enumerate}
This mechanism encourages solutions to explore novel regions of the descriptor space that are distant from other solutions.

\subsubsection{Dominated Novelty Search}
\label{sec:background-dns}
Dominated Novelty Search is a QD algorithm that implements local competition using a \textsc{competition} function that rewards solutions for being different from their fitter neighbors. Given a population with fitness values $\fitnesses$ and descriptors $\descriptors$, the $\textsc{competition}$ function operates as follows:
\begin{enumerate}
    \item For each solution $i$, compute the descriptor space distances to fitter solutions $\mathcal{D}_i = \set{\norm{\descriptor_i - \descriptor_j}}{j = 1, \dots, N, \fitness_j > \fitness_i}$.
    \item Compute $\tilde{\fitness}_i$ as the \textit{dominated novelty score}: the average distance to the $k$-nearest-fitter neighbors.
\end{enumerate}
This mechanism encourages solutions to explore regions of the descriptor space that are distant from better-performing solutions, naturally balancing quality and diversity.

\subsection{Set Operations via Dot-Product Attention}
\label{sec:background-attention}
Evolutionary algorithms fundamentally operate on populations --- unordered sets of individuals. This set-based nature of populations creates a crucial requirement: any operation performed on the population must be invariant to the ordering of individuals.

Scaled dot-product attention~\citep{vaswani_AttentionAllYou_2017} provides an elegant solution to this requirement through its inherent permutation equivariance properties.
Consider a set of $N$ vectors represented as a matrix $\mathbf{X} \in \mathbb{R}^{N \times D}$. Scaled dot-product attention projects these elements into three distinct $D_K$-dimensional latent spaces --- queries $\mathbf{Q}$, keys $\mathbf{K}$, and values $\mathbf{V}$. The output $\mathbf{Y}$ is then computed as a weighted combination of the values:
\begin{align*}
\mathbf{Q} &= \mathbf{X}\mathbf{W}_Q \in \mathbb{R}^{N \times D_K} && \nonumber\\
\mathbf{K} &= \mathbf{X}\mathbf{W}_K \in \mathbb{R}^{N \times D_K} && \mathbf{Y} = \text{softmax}\left(\frac{\mathbf{Q}\mathbf{K}^\top}{\sqrt{D_K}}\right)\mathbf{V} \in \mathbb{R}^{N \times D_K}\\
\mathbf{V} &= \mathbf{X}\mathbf{W}_V \in \mathbb{R}^{N \times D_K} && \nonumber
\end{align*}

Crucially, this transformation exhibits permutation equivariance: permuting the rows of the input population matrix $\mathbf{X}$ results in the same permutation being applied to the rows of the output $\mathbf{Y}$~\citep{vaswani_AttentionAllYou_2017}. This property makes scaled dot-product attention and transformer architectures particularly well-suited for evolutionary algorithms, as it naturally preserves the set-based structure of populations while enabling complex interactions between individuals~\citep{lange_DiscoveringEvolutionStrategies_2023,lange_DiscoveringAttentionBasedGenetic_2023,lange_EvolutionTransformerInContext_2024}.

\section{Method}
In this section, we present our approach to discovering novel Quality-Diversity algorithms through Meta-Black-Box Optimization.
First, we introduce Learned Quality-Diversity (LQD), a flexible framework that can represent any QD algorithm by parameterizing its competition function (\Cref{sec:competition}).
Then, we detail the meta-learning procedure used to discover effective LQD algorithms (\Cref{sec:meta-bbo}).

\subsection{Learned Quality-Diversity Algorithm}
\label{sec:competition}
As shown in \Cref{alg:ga,alg:qd}, Quality-Diversity algorithms can be viewed as Genetic Algorithms with local competition replacing global competition. The key difference lies in the \textsc{competition} function: while GAs use identity mapping (global competition), QD algorithms implement various forms of local competition. For instance, ME (\Cref{sec:background-me}) uses grid-based competition, while Dominated Novelty Search (\Cref{sec:background-dns}) competes with nearest fitter neighbors. Learned Quality-Diversity generalizes this framework by parameterizing the \textsc{competition} function as a transformer neural network with parameters $\theta$. \Cref{alg:learned-qd} outlines the complete LQD algorithm.

We choose the transformer architecture~\citep{vaswani_AttentionAllYou_2017} for two key properties that are crucial in our context. First, as discussed in \Cref{sec:background-attention}, transformers are permutation equivariant - the output for each individual depends on its relationship with the whole population, regardless of how individuals are ordered. This property is essential for evolutionary algorithms where the ordering of individuals in the population should not affect the competition outcomes. Second, transformers represent an architecture with minimal inductive bias, making them highly expressive and capable of representing virtually any competition rule.

\begin{algorithm}
\caption{Learned Quality-Diversity}
\label{alg:learned-qd}
\small
\begin{algorithmic}
\Require population size $N$, reproduction batch size $B$, parameters~$\theta$
\State Initialize population $\population$ with fitness $\fitnesses$ and descriptors $\descriptors$
\For{each generation}
    \State $\mathrlap{\population^\prime}\hphantom{\fitnesses, \descriptors} \gets \textsc{reproduction}(\population, \fitnesses)$\Comment{Generate $B$ offspring}
    \State $\mathrlap{\population}\hphantom{\fitnesses, \descriptors} \gets \textsc{concat}(\population, \population^\prime)$\Comment{Add offspring to population}
    \State $\mathrlap{\fitnesses, \descriptors}\hphantom{\fitnesses, \descriptors} \gets \textsc{evaluation}(X)$\Comment{Evaluate fitness and descriptor}
    \State {\color{blue}$\mathrlap{\tilde{\fitnesses}}\hphantom{\fitnesses, \descriptors} \gets \textsc{competition}_\theta(\fitnesses, \descriptors)$\Comment{Learned local competition}}
    \State $\mathrlap{\population}\hphantom{\fitnesses, \descriptors} \gets \textsc{selection}(\population, \tilde{\fitnesses})$\Comment{Keep top-$N$ individuals}
\EndFor
\end{algorithmic}
\end{algorithm}

\begin{figure}[h]
    \centering
    \includegraphics[width=\linewidth]{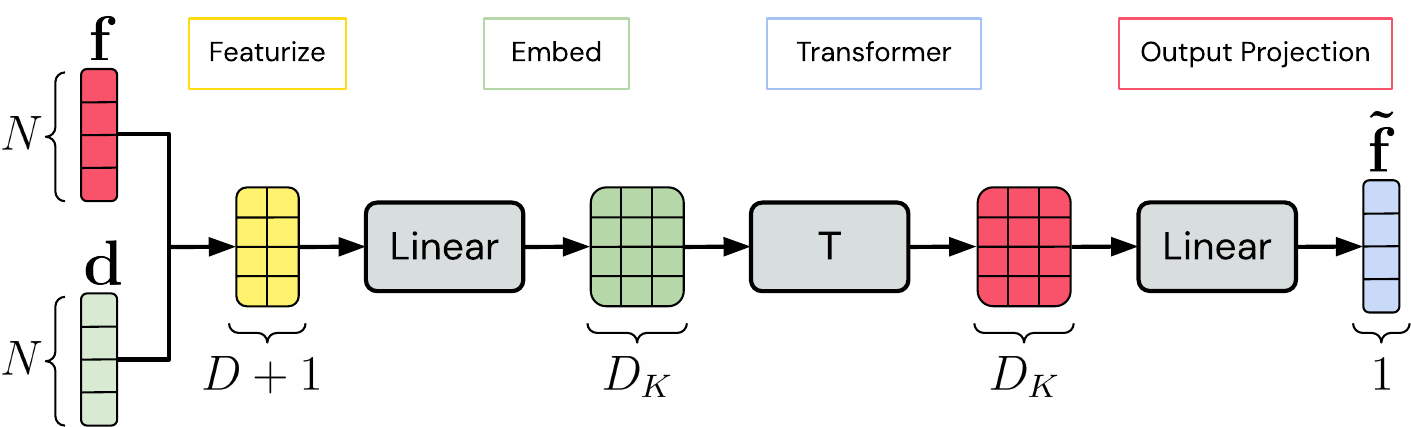}
    \caption{Learned competition function architecture.}
    \Description{Diagram showing the architecture of the learned competition function.}
    \label{fig:architecture}
\end{figure}

The learned competition function processes the population's fitness and descriptor through several transformations (\Cref{fig:architecture}):
\begin{enumerate}
	\item \textbf{Featurize}: Fitness values $\fitnesses$ and descriptors $\descriptors$ are concatenated and standardized across the population to form a matrix $\mathbf{z} \in \mathbb{R}^{N\times(D+1)}$.
	\item \textbf{Embed}: The standardized features $\mathbf{z}$ are projected into a higher-dimensional space to form embeddings suitable for transformer processing.
	\item \textbf{Transformer}: These embeddings are processed by a transformer network~\citep{vaswani_AttentionAllYou_2017} that maintains permutation equivariance, ensuring the competition rules remain consistent regardless of population ordering.
	\item \textbf{Output projection}: The transformer's output ($\mathbb{R}^{N \times D}$) is projected to scalar values ($\tilde{\fitnesses} \in \mathbb{R}^N$) that determine survival in the selection step.
\end{enumerate}

Notably, we do not embed seemingly fundamental evolutionary principles into the architecture --- even the common practice of maintaining the highest-fitness solutions in the population must be discovered if it proves beneficial. Through meta-black-box optimization, this flexible architecture allows us to explore the vast space of possible Quality-Diversity competition rules with minimal constraints, effectively searching through all archive mechanisms.


\subsection{Meta-Black-Box Optimization}
\label{sec:meta-bbo}
To discover effective LQD algorithms, we adapt the meta-learning procedure introduced by~\citet{lange_DiscoveringEvolutionStrategies_2023,lange_DiscoveringAttentionBasedGenetic_2023} for evolving evolutionary algorithms. Our meta-optimization approach consists of three key components: a diverse set of optimization problems that form the meta-task (\Cref{sec:meta-tasks}), performance metrics that define the meta-objective (\Cref{sec:meta-objective}), and an efficient meta-training procedure (\Cref{sec:meta-training}).

\subsubsection{Meta-Task}
\label{sec:meta-tasks}
We train LQD on a diverse distribution of 22 BBOB functions from~\citet{finck_RealParameterBlackBoxOptimization_noiseless}. These functions encompass a wide range of optimization challenges, including varying degrees of separability, conditioning, and multi-modality, ensuring our learned algorithm develops robust strategies. The complete set of training functions is detailed in~\Cref{tab:bbob}. To further enhance the robustness of our LQD algorithm, we incorporate different noise models from~\citet{finck_RealParameterBlackBoxOptimization_noisy} during training, including uniform, Gaussian, and Cauchy noise distributions. This noise injection helps ensure the learned competition rules remain effective even under noisy fitness evaluations.

During meta-training, we employ extensive data augmentation on the BBO tasks to improve generalization. Specifically, BBOB functions are sampled with varying dimensionality (from 2 to 12 dimensions), different noise model parameters, and random rotations of the search space. This augmentation strategy creates a rich training distribution that helps the learned algorithm develop robust competition rules that generalize across problems.

Finally, to transform these BBO tasks into QD tasks, we associate each task with a descriptor space through random projection. When sampling a task during meta-training, we generate a random matrix $\mathbf{N} \in \mathbb{R}^{D \times n}$ with independent standard normal entries and compute descriptors for each solution $i$ as $\descriptor_i = \mathbf{N} \individual_i$. This approach is theoretically justified by the Johnson-Lindenstrauss lemma~\citep{Johnson1984ExtensionsOL}, which guarantees that random projections approximately preserve pairwise distances between points with high probability. By projecting high-dimensional genotypes into a lower-dimensional descriptor space while maintaining their relative distances, we create meaningful characteristics that capture solution similarities without requiring domain-specific descriptors. This domain-agnostic approach allows us to evaluate our method across diverse optimization problems while ensuring the descriptor space remains informative for local competition.

\subsubsection{Meta-Objective}
\label{sec:meta-objective}
To demonstrate the flexibility of our approach, we explore three distinct meta-objectives, each giving rise to a specialized LQD variant:
\begin{itemize}
    \item \textbf{Fitness}: Optimizing the highest fitness value in the population gives rise to LQD (F), designed for tasks where peak performance is the primary goal.
    \item \textbf{Novelty}: Maximizing the average novelty score across the population, as defined in~\Cref{sec:background-ns}, leads to LQD (N), suited for exploration-focused tasks.
    \item \textbf{QD score}: Balancing both fitness and diversity produces LQD (F+N), designed for illumination. The QD score is defined as the maximum fitness multipled by the average dominated novelty score.
\end{itemize}

\subsubsection{Meta-Training}
\label{sec:meta-training}
We optimize the LQD parameters on the diverse family of BBO tasks described in~\Cref{sec:meta-tasks}, following the Meta-BBO procedure established in previous work~\citep{lange_DiscoveringEvolutionStrategies_2023,lange_DiscoveringAttentionBasedGenetic_2023,lange_EvolutionTransformerInContext_2024}. Each set of parameters $\theta_i$ defines a unique QD algorithm by specifying how solutions compete with each other through different competition functions. The meta-training process, illustrated in~\Cref{fig:teaser}, maintains a meta-population of $M=256$ different LQD parameter sets, denoted as $\theta_i$ for $i=1,...,M$. At each meta-generation, we uniformly sample a set of $K=256$ BBO tasks and evaluate each LQD variant across all tasks. These evaluations are aggregated into meta-objectives that guide the meta-evolutionary optimization.

The LQD parameters are trained for 16,384 meta-generations using Sep-CMA-ES, with each inner loop running for 256 generations with a population size of 128. After meta-training, we select the best-performing parameter set based on validation performance across held-out tasks, resulting in three specialized LQD variants (F, N, and F+N). Through this process, we effectively search through the space of possible Quality-Diversity algorithms to discover competition rules that are optimized for different objectives.

\begin{algorithm}[h]
\caption{Meta-Black-Box Optimization of Learned QD}
\label{algo:meta-bbo}
\small
\begin{algorithmic}
\Require Meta-ES with meta-population size $M$, meta-task with meta-batch size $K$, population size $N$, reproduction batch size $B$
\State Initialize meta-ES distribution $(\mu$, $\Sigma)$
\While{not done}
    \State Sample $K$ BBO tasks $\xi_k, \forall k = 1, \dots, K$
    \State Sample $M$ LQD parameters: $\theta_i \sim \mathcal{N}(\mu, \Sigma), \forall i = 1, \dots, M$
    \State Evaluate all LQD individuals on the same $K$ tasks:
    \For{$k = 1, \dots, K$}
        \For{$i = 1, \dots, M$}
        \State Inner loop (\Cref{alg:learned-qd}) to get $\left[(\individual_j)_{j=1}^N\right]_{t=1}^T$
        \EndFor
    \EndFor
    \State Collect fitness scores $\left[\left[\left[\left(f\left(\individual_{j, t} \middle| \xi_k\right)\right)_{j=1}^N\right]_{t=1}^T\right]_{k=1}^K \middle| \theta_i \right]_{i=1}^M$
    \State Compute normalized and aggregated meta-fitness $(\tilde{f}(\theta_i))_{i=1}^M$
    \State Update distribution $(\mu, \Sigma) \leftarrow \text{Meta-ES}\left((\tilde{f}(\theta_i))_{i=1}^M \middle| \mu, \Sigma\right)$
\EndWhile
\end{algorithmic}
\end{algorithm}

The complete meta-optimization process is detailed in~\Cref{algo:meta-bbo}, with comprehensive hyperparameter settings provided in~\Cref{tab:hyperparameters}. Notably, the transformer architecture uses 4 layers with 16 features and 4 attention heads, striking a balance between expressivity and computational efficiency.

\section{Experiments}
We conduct an extensive empirical evaluation of our meta-optimization procedure and the discovered Learned QD algorithms. Our experiments are designed to answer three key research questions:
\begin{itemize}
    \item Can we meta-evolve QD algorithms that demonstrate competitive performance against established baselines on both meta-training tasks and out-of-distribution BBOB functions on different objectives (\Cref{sec:experiment-bbo})?
    \item How well do these learned algorithms generalize to fundamentally different domains like robot control tasks, which feature higher dimensionality and domain-specific behavioral descriptors (\Cref{sec:experiment-robot})?
    \item What competition strategies emerge when optimizing different meta-objectives, and how do these learned mechanisms relate to existing approaches (\Cref{sec:experiment-analysis})?
\end{itemize}
For statistical rigor, we conduct 32 independent replications of each experiment using distinct random seeds. Statistical significance is assessed using the Wilcoxon–Mann–Whitney $U$ test with Holm-Bonferroni correction. Our implementation leverages hardware acceleration through JAX~\citep{jax}, with optimization algorithms from evosax~\citep{evosax2022github} and QDax~\citep{qdax}. All experiments were conducted on 8 H100 GPUs.

\subsection{Baselines}
\label{sec:baselines}
We compare LQD against four established baseline algorithms:
\begin{itemize}
    \item MAP-Elites: A QD algorithm that partitions solutions into grid cells for local competition.
    \item Dominated Novelty Search: A QD algorithm using nearest-neighbor local competition.
    \item Genetic Algorithm: A traditional evolutionary algorithm using global competition.
    \item Novelty Search: A variant of novelty search where novelty scores are computed using only the current population.
    \item Random: A GA variant where the competition function assigns random fitness values, resulting in random selection.
\end{itemize}

\subsection{Black-Box Optimization Benchmark Tasks}
\label{sec:experiment-bbo}
We conducted extensive experiments to evaluate the performance of our three LQD variants (F, N, and F+N) against established baselines across both the meta-training tasks (\Cref{tab:bbob}) and a set of challenging out-of-distribution optimization problems (\Cref{tab:bbob-ood}). These three LQD models, trained once, were used without any task-specific fine-tuning across all experiments. For out-of-distribution BBO evaluation, we selected six complex functions: Gallagher's Gaussian 101-me Peaks Function and Gallagher's Gaussian 21-hi Peaks Function from~\citet{finck_RealParameterBlackBoxOptimization_noiseless}, as well as the Ackley, Dixon-Price, Salomon and Levy functions from~\citet{jamil_LiteratureSurveyBenchmark_2013}.

\begin{figure}[h]
    \centering
    \includegraphics[width=\linewidth]{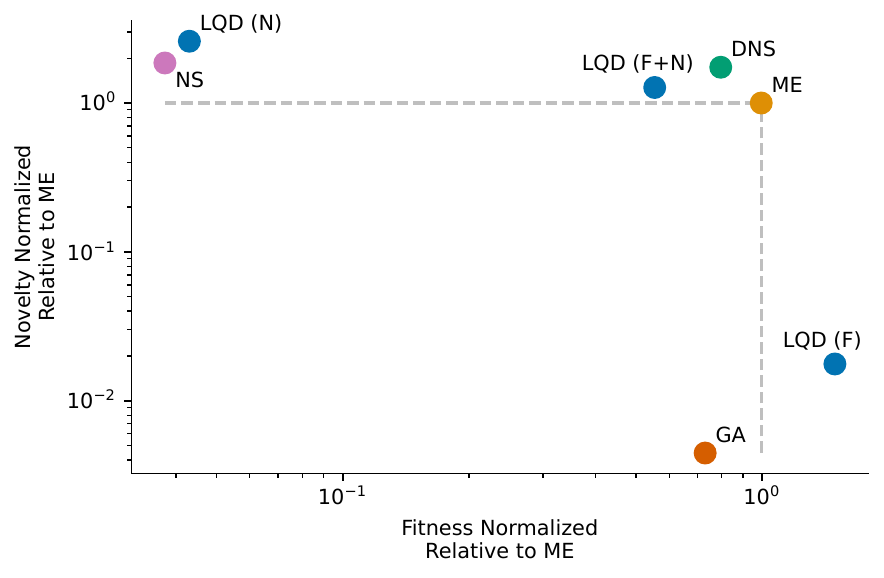}
    \caption{Quality-Diversity trade-off across algorithms. Each point represents the median fitness and novelty scores achieved across 32 runs on BBO training tasks. LQD variants demonstrate the flexibility of our framework, successfully specializing for different objectives while matching or exceeding baseline performance.}
    \Description{Pareto front for two metrics fitness and novelty for each algorithm LQD (N), LQD (F), LQD (F+N), ME, DNS, GA and NS on BBO training tasks. We can see that LQD (N) outperforms NS in novelty and LQD (F) outperforms GA in fitness. LQD (F+N) outperforms ME in novelty but not in fitness and is outperformed by DNS. However, LQD (F+N) is very close in performance to DNS and ME. This plot shows that our framework enable to discover QD algorithms effectively and enable to push in all direction (fitness only, novelty only or fitness and novelty.}
    \label{fig:pareto}
\end{figure}

Our experiments reveal several key findings about the effectiveness of LQD. First, we examined how different meta-objectives shape algorithm behavior by training three LQD variants: LQD (F) optimized for fitness, LQD (N) optimized for novelty, and LQD (F+N) optimized for QD (\Cref{sec:meta-objective}). \Cref{fig:pareto} shows the Pareto front of fitness versus novelty scores achieved by each algorithm on the training tasks. The results demonstrate that our framework successfully discovers specialized algorithms: LQD (N) achieves superior novelty scores compared to Novelty Search ($p < 10^{-9}$), while LQD (F) outperforms the standard GA in fitness optimization ($p < 10^{-10}$). LQD (F+N) approaches but does not quite match DNS's performance, while outperforming ME in novelty ($p < 10^{-7}$) but showing lower fitness scores, suggesting that simultaneously optimizing both objectives remains challenging.

\begin{figure}[h]
    \centering
    \includegraphics[width=\linewidth]{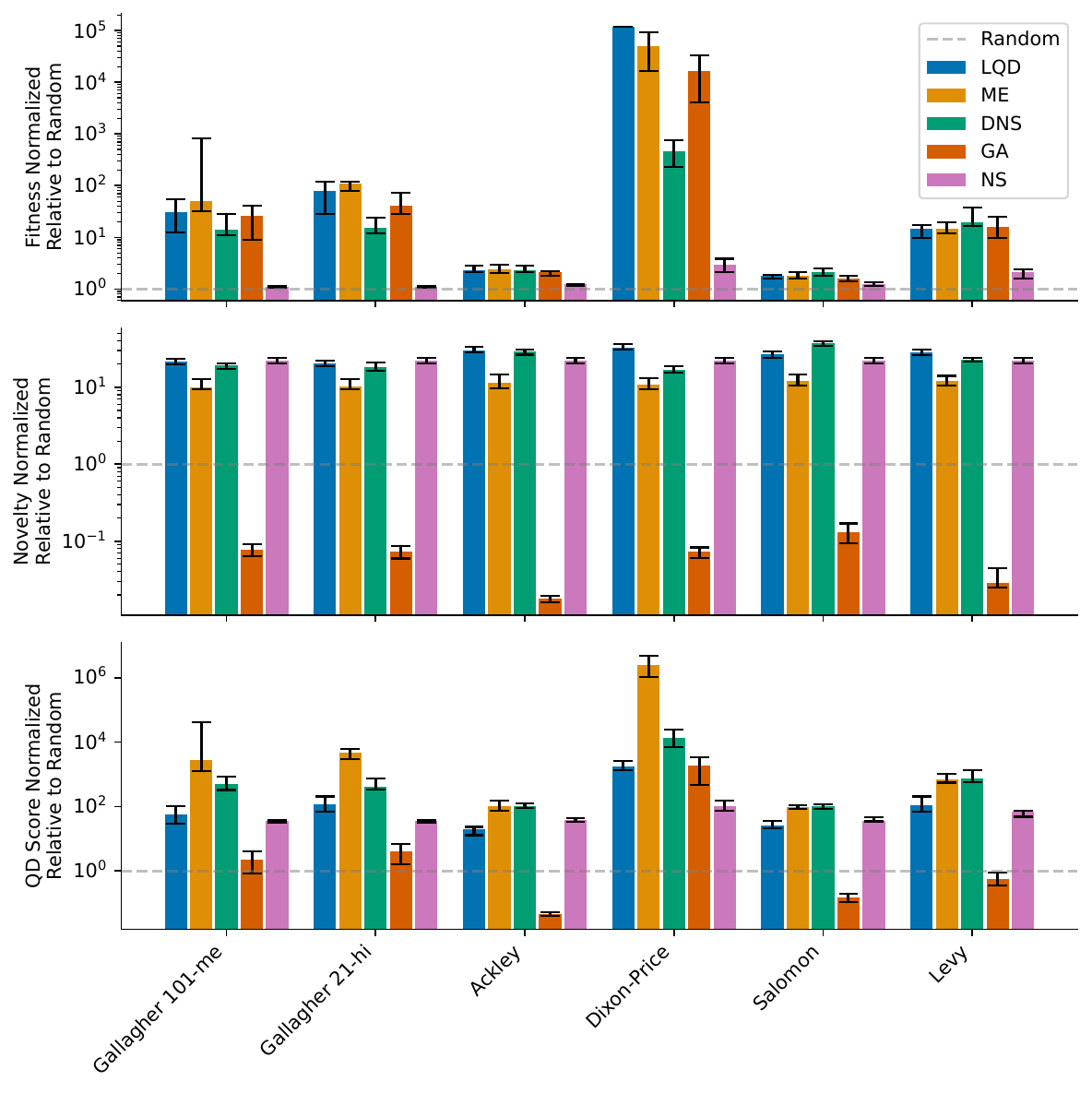}
    \caption{Performance comparison across out-of-distribution BBOB tasks for the three objectives. Results are normalized relative to random GA baseline (dashed line at y=1). Bars show median values across 32 runs, with error bars indicating interquartile range.}
    \Description{A plot wit three bar subplots comparing evolutionary algorithms on BBOB out-of-distributions tasks on the three objectives. The y-axis shows performance values on a logarithmic scale, normalized to random GA performance (shown as a horizontal dashed line at y=1). Each group of bars represents a different benchmark function, with different colors representing different algorithms. Error bars extend from the 25th to 75th percentiles of the results.}
    \label{fig:bars}
\end{figure}

The generalization capabilities of LQD are particularly evident in~\Cref{fig:bars}, which compares performance across out-of-distribution tasks. Detailed results on training tasks are available in~\Cref{fig:bars-appendix}. We normalize results relative to random GA performance to enable meaningful comparisons across different functions. LQD (N) demonstrates remarkable generalization, achieving the best novelty scores on nearly all out-of-distribution tasks, with DNS showing marginally better performance only on the Salomon function. LQD (F) maintains competitive fitness performance across these unseen problems, while LQD (F+N) shows comparable but slightly lower performance than DNS. Notably, while ME appears to achieve higher novelty scores in some cases, this is partially due to its ability to maintain partially-filled populations, which can artificially inflate novelty scores.

\begin{figure}[h]
\centering
    \includegraphics[width=\linewidth]{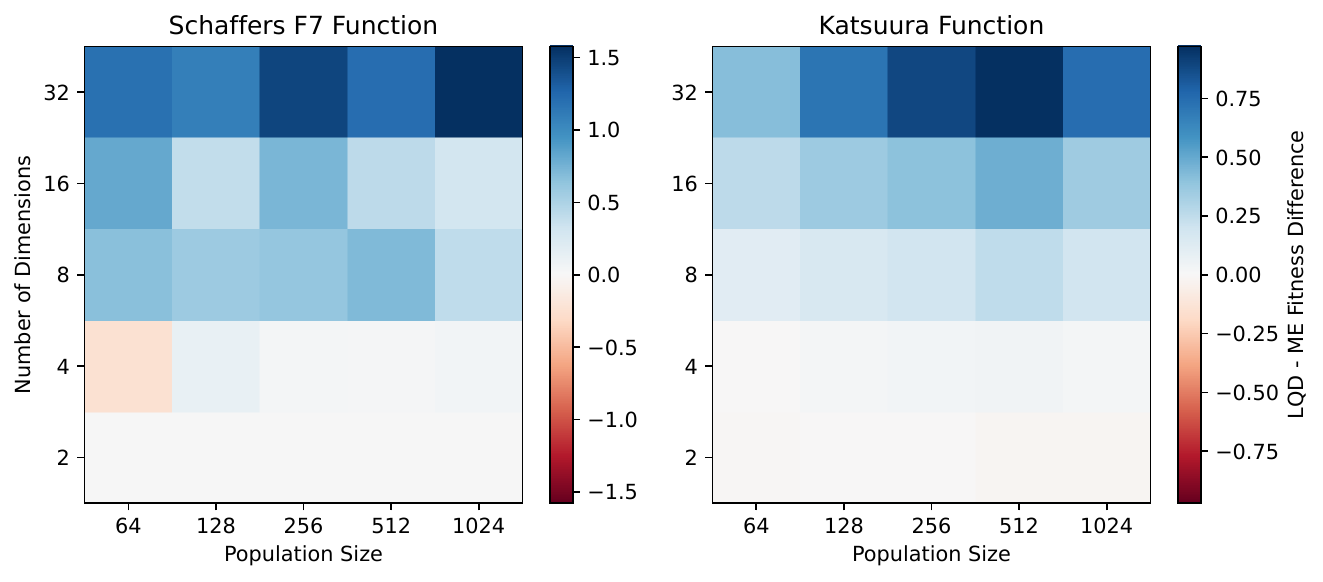}
    \caption{Generalization analysis across population sizes and search space dimensions. Heatmaps show the performance difference between LQD and ME (blue indicates LQD advantage, red indicates ME advantage) for two BBOB tasks.}
    \Description{Heatmap showing the difference in fitness between LQD and ME for different population sizes 64, 128, 256, 512, 1024 and number of dimensions 2, 4, 8, 16, 32. The heatmap are shown for 4 different BBO functions: Rastrigin Function, Linear Slope, Schaffers F7 Function and Katsuura Function.}
    \label{fig:grid}
\end{figure}

Perhaps most impressively, LQD demonstrates robust scaling properties across different population sizes and problem dimensions, as shown in~\Cref{fig:grid}. LQD (F) outperforms ME across almost all configurations, with the performance gap widening as both population size and dimensionality increase. The only exception is the Schaffers F7 function with population size 64 and dimension 4, where ME maintains a slight edge. This scaling behavior is particularly remarkable as our experiments show that LQD can maintain high performance with populations and dimensions several times larger than those used during training -- despite being trained only on population sizes of 128 and dimensions between 2 and 12. This suggests that the learned competition rules capture fundamental principles that generalize far beyond their training conditions.

\subsection{Robot Control Tasks}
\label{sec:experiment-robot}
To evaluate generalization capabilities beyond abstract benchmarks, we tested LQD (F) and baseline approaches on a suite of challenging robot control tasks. These experiments represent a significant departure from the meta-training conditions, featuring much higher-dimensional search spaces (up to 584 parameters for the Ant robot), domain-specific behavioral descriptors based on foot contact patterns and velocities (rather than random projections), and fundamentally different fitness landscapes.

\begin{figure}[h]
    \centering
    \includegraphics[width=\linewidth]{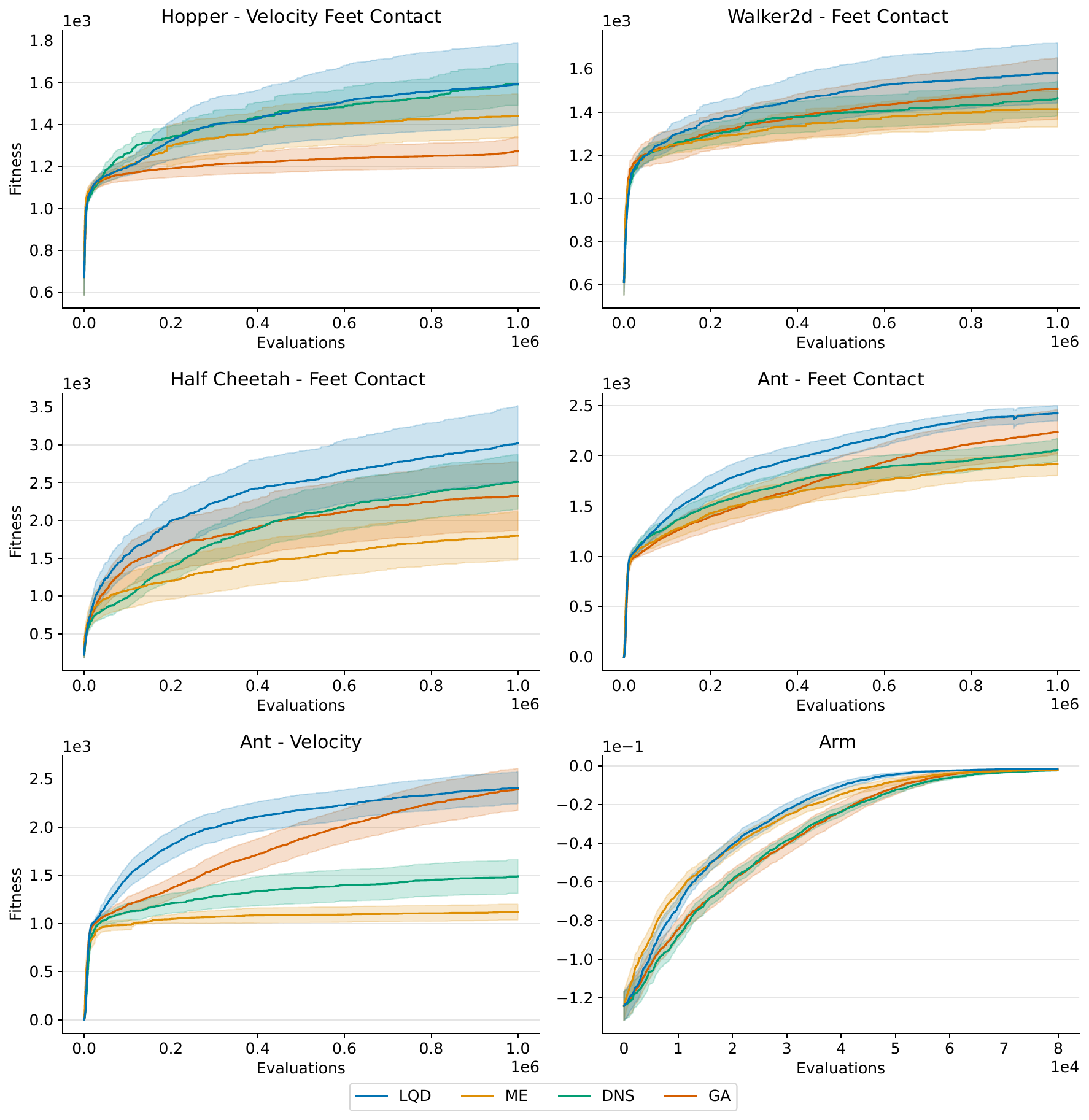}
    \caption{Fitness across robot control tasks. Lines show mean performance across 32 independent runs, with shaded regions indicating 95\% confidence intervals. LQD demonstrates strong generalization, matching or exceeding baseline performance despite never being trained on robotic tasks.}
    \Description{Plots showing fitness evolution over generations for LQD and baselines across six robot control tasks, with population size 128 and batch size 32. Results show LQD matching DNS performance on Hopper (velocity-feet), achieving best performance on Walker2d, Half Cheetah, and Ant (feet contact), matching GA's leading performance on Ant (velocity), and demonstrating fastest convergence on the arm reaching task while achieving equivalent final performance.}
    \label{fig:evo-robot}
\end{figure}

We evaluated performance across Hopper, Walker2d, Half Cheetah, Ant and Arm robots with various descriptor configurations with a budget of 1 million evaluations. As shown in~\Cref{fig:evo-robot},  LQD demonstrates remarkable generalization to these challenging domains: it matches DNS's leading performance on the Hopper task, outperforms all baselines on Walker2d, Half Cheetah, and Ant with feet contact descriptors, and equals GA's top performance on Ant with velocity-based descriptors. On the arm reaching task, LQD achieves the fastest convergence while matching the final performance of other algorithms. When aggregating results across all robot control tasks (\Cref{fig:teaser}), LQD significantly outperforms all baselines ($p < 0.005$) except GA, which it outperforms but not with statistical significance.

These results are particularly notable given that LQD was never exposed to robotic control problems, high-dimensional search spaces, or domain-specific descriptors during training, suggesting the learned competition rules capture fundamental principles that generalize well beyond their training distribution.

\section{Analysis of Discovered LQD}
\label{sec:experiment-analysis}
After demonstrating LQD's effectiveness across various optimization tasks, we now investigate the key mechanisms behind its performance. We examine three key aspects: the emergence of population diversity even when optimizing solely for fitness (\Cref{sec:emergent-diversity}), the local competition strategies discovered for different objectives (\Cref{sec:local-competition}), and the role of descriptors in identifying promising stepping stones (\Cref{sec:ablation}).

\subsection{Emergent Diversity}
\label{sec:emergent-diversity}
\begin{figure}[h]
    \centering
    \includegraphics[width=\linewidth]{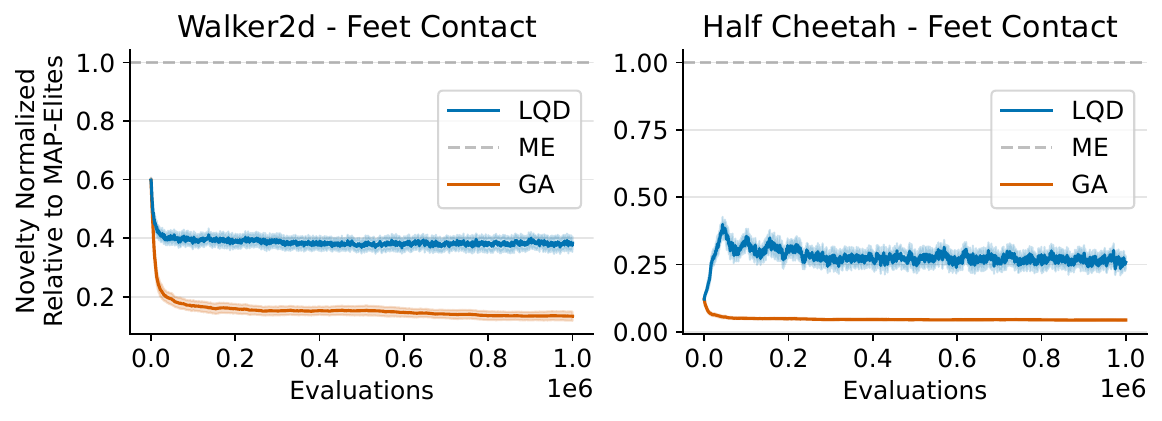}
    \caption{Novelty across robot control tasks. Lines show mean performance across 32 independent runs, with shaded regions indicating 95\% confidence intervals.}
    \Description{Novelty normalized by ME's final novelty for LQD and GA on Walker2d Feet Contact and Half Cheetah Feet Contact. We see that local competition learned by LQD is consistently fostering more diversity than a pure GA (LQD is around 40 percent of ME while GA is below 10 percent). It is interesting because meta-learning shows that diversity is important. We rediscovered that keeping exploring and finding interesting stepping is important to achieving good fitness.}
    \label{fig:emergent-diversity}
\end{figure}

A striking finding from our analysis is that even LQD variants trained purely for fitness optimization naturally maintain significant population diversity. As shown in~\Cref{fig:pareto}, while LQD (F) outperforms GA in maximizing fitness, it simultaneously achieves substantially higher novelty scores despite never being trained for diversity. This emergent diversity is further demonstrated in robot control tasks (\Cref{fig:emergent-diversity}). For example, on Walker2d, LQD (F) consistently achieves novelty scores approximately 40\% of ME's final diversity, while standard GA maintains less than 10\% of ME's diversity, demonstrating much stronger convergence.

This emergent diversity in LQD suggests that meta-optimization has rediscovered a fundamental principle of evolutionary systems: maintaining a diverse population creates crucial stepping stones that enable the discovery of high-performing solutions. While traditional GAs often converge prematurely due to their global selection, LQD appears to have learned more sophisticated competition dynamics that naturally preserve promising intermediate solutions.

\subsection{Discovered Local Competition Strategies}
\label{sec:local-competition}
\begin{figure}[h]
    \centering
    \includegraphics[width=\linewidth]{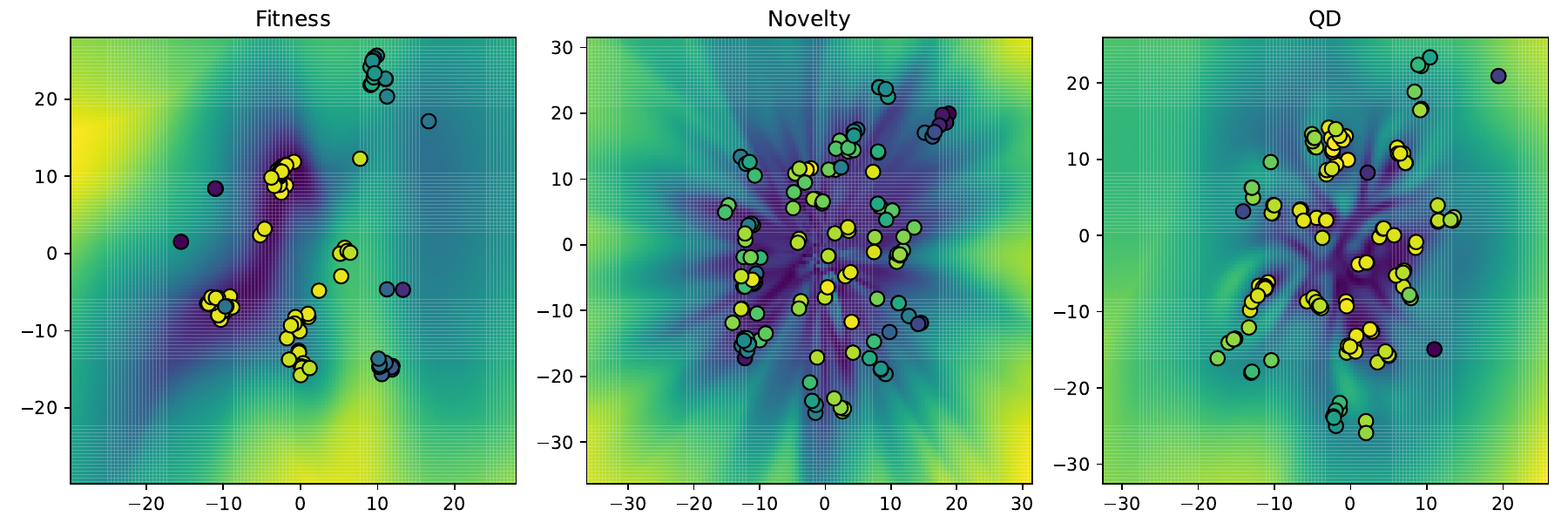}
    \caption{Visualization of learned competition landscapes across LQD variants. Points show evolved populations colored by raw fitness, while heatmaps represent the competition fitness $\tilde{f}$ that would be assigned to a solution with median fitness at each location.}
    \label{fig:heatmaps}
\end{figure}

To understand how different meta-objectives shape the competition rules learned by LQD, we visualize the competition fitness values $\tilde{f}$ assigned by each variant across the descriptor space. Figure~\ref{fig:heatmaps} shows these learned competition landscapes, where heatmaps represent the competition fitness that would be assigned to a solution with median raw fitness at each point in the descriptor space, based on the current state of the population (represented as colored dots).

LQD (N) develops a simple distance-based competition mechanism that rewards solutions for being far from existing ones, effectively rediscovering the core principle of novelty search. LQD (F), despite being trained purely for fitness, learns a more nuanced strategy that creates fitness-sensitive reward patterns around promising solutions, explaining the emergent diversity observed in Section~\ref{sec:emergent-diversity}. LQD (F+N) combines these approaches, creating directed exploration through competition rules that balance novelty seeking with fitness sensitivity.

These visualizations reveal that meta-optimization discovers qualitatively different competition strategies based on the meta-objective, from pure novelty-seeking to sophisticated hybrid mechanisms. Notably, even when optimizing for fitness alone, LQD learns to maintain strategic diversity through local competition, suggesting population diversity emerges as an instrumental goal for achieving high performance.

\vspace{-0.5cm}
\subsection{Impact of Descriptors on LQD}
\label{sec:ablation}
To understand if LQD leverages descriptor information to guide optimization, we conducted an ablation study comparing three configurations: standard LQD with task-specific descriptors, LQD with random descriptors (where descriptors are sampled from a standard normal distribution, breaking any relationship with the underlying solutions), and a standard GA without descriptors.

\begin{figure}[h]
\centering
\includegraphics[width=\linewidth]{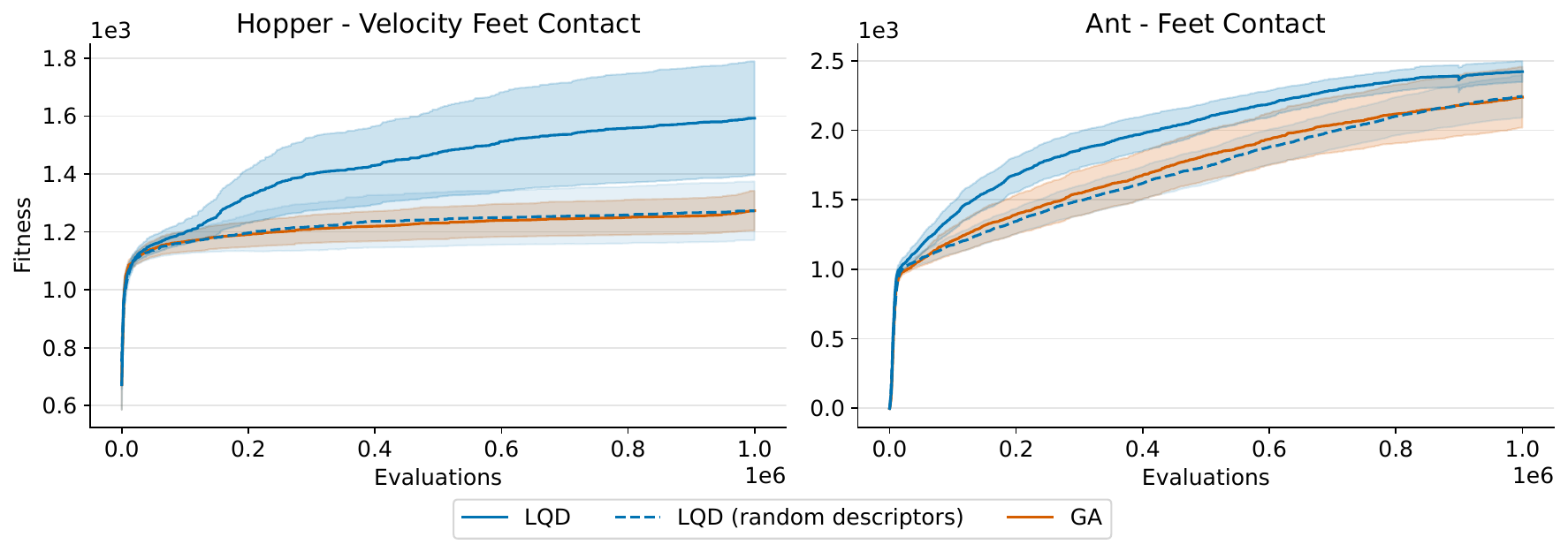}
    \caption{Impact of descriptors on optimization performance. Comparison of LQD with task-specific descriptors, LQD with random descriptors, and standard GA on robot control tasks. Lines show mean performance across 32 independent runs with 95\% confidence intervals. The convergence of LQD with random descriptors to GA performance demonstrates that meaningful descriptor information is crucial for LQD's enhanced optimization capabilities.}
    \Description{Fitness comparison plots showing LQD consistently outperforming both LQD with random descriptors and GA across robot control tasks. LQD with random descriptors performs similarly to GA, indicating that the algorithm requires meaningful descriptor information to achieve superior performance.}
    \label{fig:ablation}
\end{figure}

As shown in~\Cref{fig:ablation}, LQD with random descriptors performs similarly to GA across robot control tasks, suggesting that when descriptor information is meaningless, LQD essentially defaults to global competition. In contrast, LQD with task-specific descriptors significantly outperforms both alternatives, demonstrating that the learned competition rules effectively exploit the structure encoded in the descriptor space to identify promising stepping stones.

This performance gap reveals that LQD's success stems from its ability to leverage meaningful relationships between solutions captured by the descriptor space, rather than from the mere presence of additional input features. The algorithm appears to have discovered that descriptor-based similarity can indicate promising search directions, using this information to maintain strategic diversity that facilitates the discovery of high-performing solutions.

\vspace{-0.5cm}
\section{Conclusion}
Our work demonstrates that meta-learning can discover sophisticated Quality-Diversity algorithms that outperform traditional hand-designed approaches. Our Learned Quality-Diversity framework, which parameterizes competition rules using attention-based architectures, successfully discovers algorithms that capture complex relationships between solutions while maintaining the benefits of local competition that make biological evolution so powerful.

The discovered algorithms show several remarkable properties. First, they achieve superior or competitive performance compared to established baselines across various optimization tasks. Second, they demonstrate robust generalization, performing well on problems far outside their training distribution. Third, and perhaps most intriguingly, even when trained solely for fitness optimization, they naturally maintain significant population diversity, suggesting that meta-learning rediscovers diversity as an instrumental goal for achieving peak performance.

Our results also highlight areas for further improvement. While LQD variants excel at pure fitness maximization or novelty seeking, simultaneously optimizing both objectives remains challenging. LQD (F+N) approaches but does not consistently outperform existing QD algorithms, indicating that balancing quality and diversity remains a fundamental challenge in evolutionary computation.


\bibliographystyle{ACM-Reference-Format}
\bibliography{ref}

\clearpage
\appendix

\section*{Supplementary Materials}
\section{Additional Results}
\subsection{Black-Box Optimization Benchmark Tasks}
\begin{figure*}
    \centering
    \includegraphics[width=\linewidth]{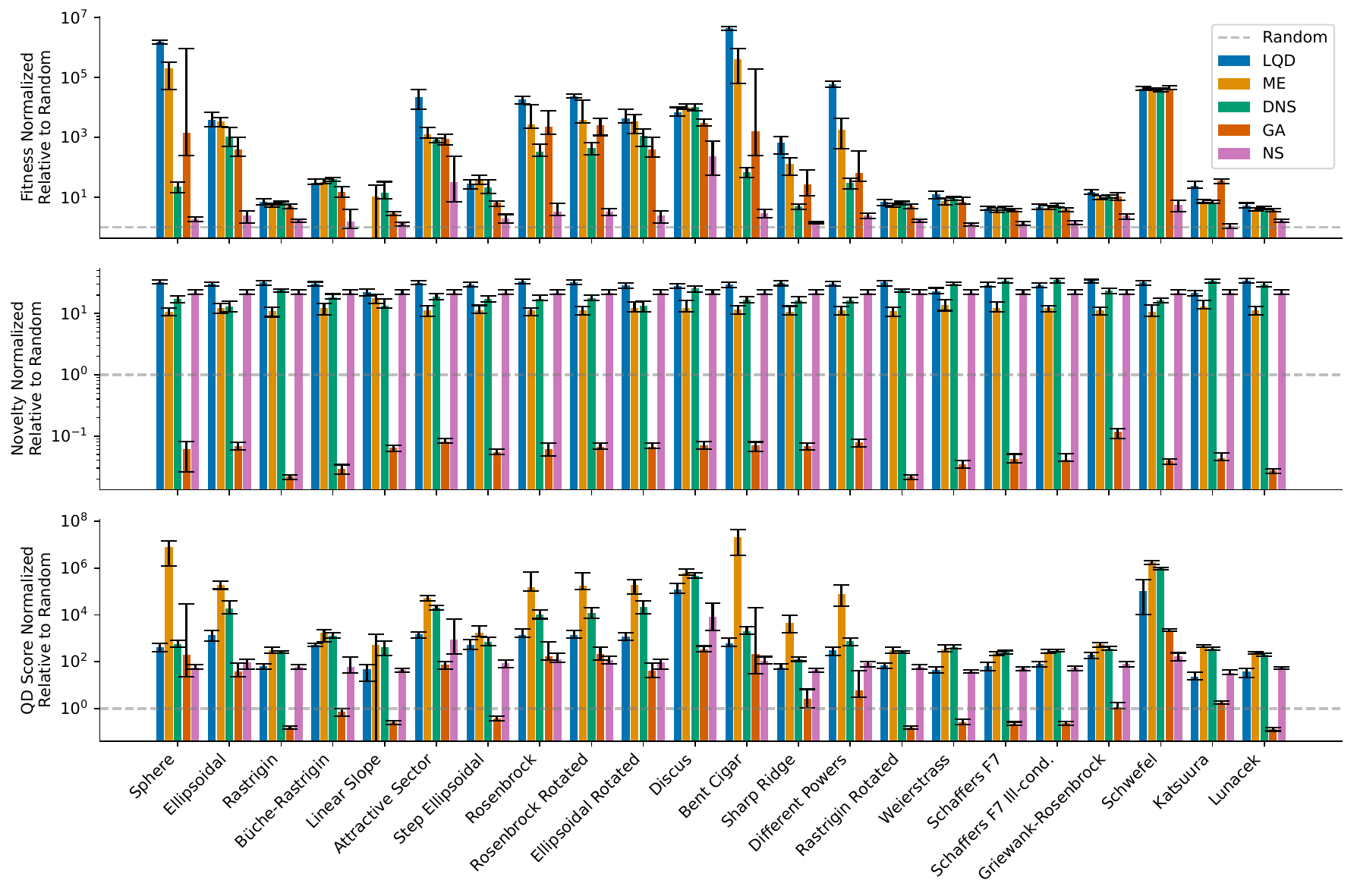}
    \caption{Performance comparison across meta-training BBOB tasks for the three distinct objectives. Results are normalized relative to random GA baseline (dashed line at y=1). Bars show median values across 32 runs, with error bars indicating interquartile range. Higher values indicate better performance.}
    \Description{}
    \label{fig:bars-appendix}
\end{figure*}

To complement the out-of-distribution results presented in Section~\ref{sec:experiment-bbo}, we provide comprehensive performance comparisons across all meta-training BBOB tasks. \Cref{fig:bars-appendix} shows normalized performance metrics for each algorithm variant on the training functions described in~\Cref{appendix:bbob}. The results reinforce our main findings: LQD (N) consistently achieves superior novelty scores, while LQD (F) maintains competitive performance in fitness optimization.

\subsection{Impact of Descriptors on LQD}
\begin{figure}[H]
    \centering
    \includegraphics[width=\columnwidth]{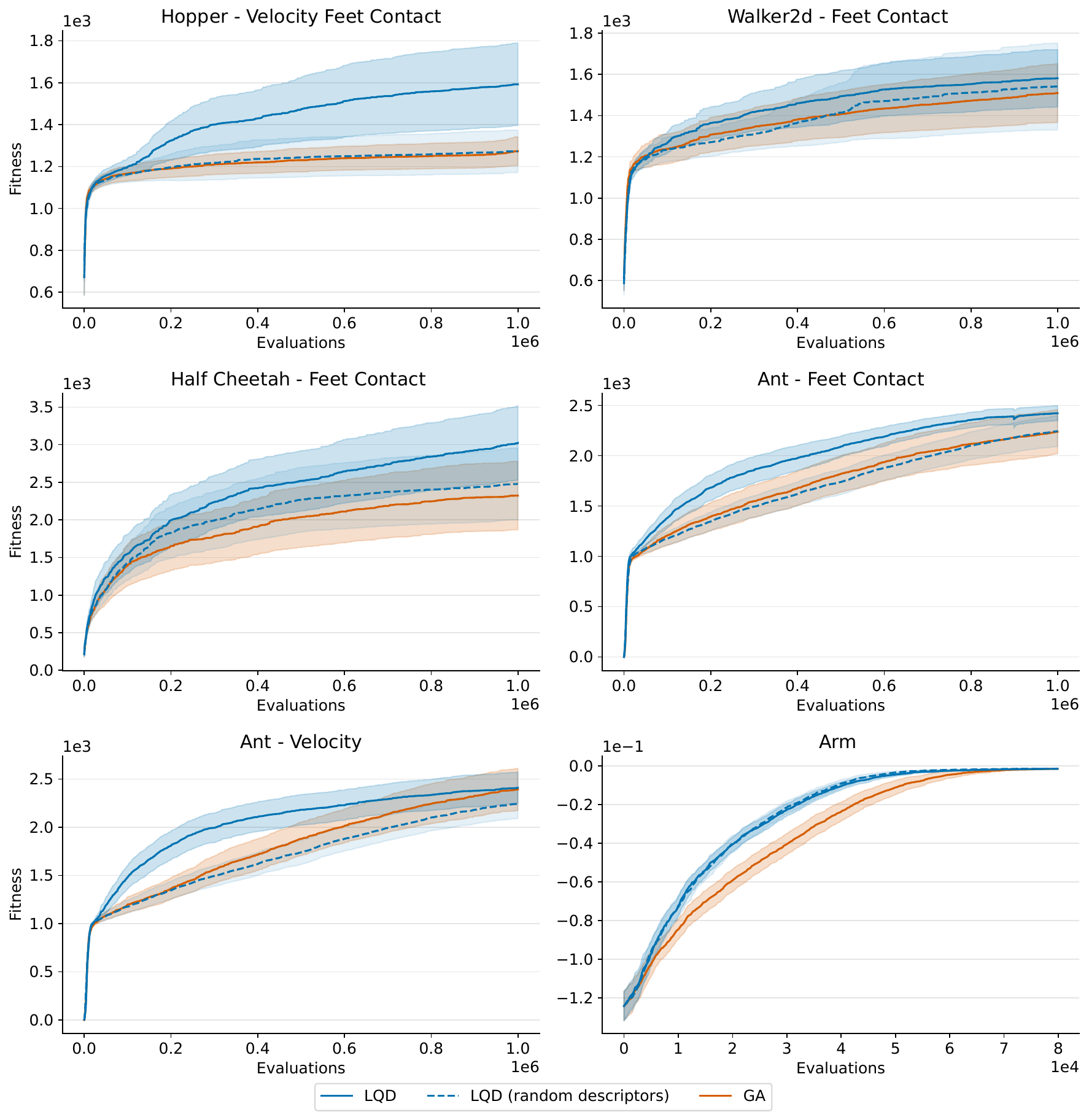}
    \caption{Impact of descriptors on optimization performance. Comparison of LQD with task-specific descriptors, LQD with random descriptors, and standard GA on robot control tasks. Lines show mean performance across 32 independent runs with 95\% confidence intervals. The convergence of LQD with random descriptors to GA performance demonstrates that meaningful descriptor information is crucial for LQD's enhanced optimization capabilities.}
    \Description{}
    \label{fig:ablation-appendix}
\end{figure}

Building on the ablation analysis in~\Cref{sec:ablation},~\Cref{fig:ablation-appendix} presents additional robot control tasks that further validate our findings regarding the importance of meaningful descriptors. The results consistently show that LQD with random descriptors performs comparably to standard GA, while LQD with task-specific descriptors maintains superior performance. This pattern holds across different robot morphologies and descriptor types, providing strong evidence that LQD's enhanced performance stems from its ability to leverage meaningful relationships encoded in the descriptor space rather than from increased model capacity alone.

\section{Hyperparameters}
Our meta-optimization framework involves three sets of hyperparameters, detailed in \Cref{tab:hyperparameters}. For the outer loop meta-optimization, we use Separable CMA-ES with an initial step size $\sigma=0.1$. The meta-population size ($M=256$) and meta-batch size ($K=256$) were chosen to provide stable gradient estimates while remaining computationally tractable. We run the meta-optimization for 16,384 generations to ensure convergence of the learned algorithms.

For the inner loop optimization, we use relatively modest population sizes ($N=128$) and reproduction batch sizes ($B=32$). Each inner loop runs for 256 generations, providing sufficient time for the algorithms to demonstrate their optimization capabilities while keeping meta-training computationally feasible.

The transformer architecture hyperparameters were selected to balance expressiveness and computational efficiency. We use 4 attention layers with 16 features per layer and 4 attention heads, resulting in approximately 5,000 trainable parameters. This architecture provides sufficient capacity to learn sophisticated competition rules while remaining small enough to meta-optimize effectively.
\begin{table}[H]
\caption{Meta-black-box optimization hyperparameters}
\label{tab:hyperparameters}
\begin{tabular}{ll}
    \toprule
    Parameter & Value\\
    \midrule
    Meta-ES & Sep-CMA-ES\\
    $\sigma$ init & $0.1$\\
    Meta-population size $M$ & 256\\
    Meta-batch size $K$ & 256\\
    Num. meta-generations & 16,384\\
    \midrule
    Population size $N$ & 128\\
    Reproduction batch size $B$ & 32\\
    Num. generations $T$ & 256\\
    \midrule
    Num. layers & 4\\
    Num. features $D_K$ & 16\\
    Num. heads & 4\\
    \bottomrule
\end{tabular}
\end{table}

\section{Tasks}
\subsection{Meta-Black-Box Optimization Tasks}
\label{appendix:bbob}
For meta-optimization, we utilize 22 functions from the Black-Box Optimization Benchmark (BBOB) suite~\citep{finck_RealParameterBlackBoxOptimization_noiseless}. These functions were carefully selected to represent diverse optimization challenges across five categories: separable functions, low/moderate conditioning, high conditioning with unimodality, multimodal functions with adequate global structure, and multimodal functions with weak global structure. We exclude two computationally intensive functions (Gallagher's Gaussian 101-me Peaks Function~\citep[p.~105]{finck_RealParameterBlackBoxOptimization_noiseless} and 21-hi Peaks Function~\citep[p.~110]{finck_RealParameterBlackBoxOptimization_noiseless}) from the training set, reserving them instead for out-of-distribution evaluation. \Cref{tab:bbob} provides a comprehensive overview of the training functions and their properties.
\begin{table*}
\caption{Meta-black-box optimization tasks}
\label{tab:bbob}
\begin{tabular*}{\linewidth}{@{\extracolsep{\fill}} lll }
    \toprule
    Function & Property & Reference\\
    \midrule
    Sphere Function & Separable & \citet[p.~5]{finck_RealParameterBlackBoxOptimization_noiseless}\\
    Ellipsoidal Function & Separable & \citet[p.~10]{finck_RealParameterBlackBoxOptimization_noiseless}\\
    Rastrigin Function & Separable & \citet[p.~15]{finck_RealParameterBlackBoxOptimization_noiseless}\\
    Büche-Rastrigin Function & Separable & \citet[p.~20]{finck_RealParameterBlackBoxOptimization_noiseless}\\
    Linear Slope & Separable & \citet[p.~25]{finck_RealParameterBlackBoxOptimization_noiseless}\\
    \midrule
    Attractive Sector Function & Low or moderate conditioning & \citet[p.~30]{finck_RealParameterBlackBoxOptimization_noiseless}\\
    Step Ellipsoidal Function & Low or moderate conditioning & \citet[p.~35]{finck_RealParameterBlackBoxOptimization_noiseless}\\
    Rosenbrock Function & Low or moderate conditioning & \citet[p.~40]{finck_RealParameterBlackBoxOptimization_noiseless}\\
    Rosenbrock Function, rotated & Low or moderate conditioning & \citet[p.~45]{finck_RealParameterBlackBoxOptimization_noiseless}\\
    \midrule
    Ellipsoidal Function, rotated & High conditioning and unimodal & \citet[p.~50]{finck_RealParameterBlackBoxOptimization_noiseless}\\
    Discus Function & High conditioning and unimodal & \citet[p.~55]{finck_RealParameterBlackBoxOptimization_noiseless}\\
    Bent Cigar Function & High conditioning and unimodal & \citet[p.~60]{finck_RealParameterBlackBoxOptimization_noiseless}\\
    Sharp Ridge Function & High conditioning and unimodal & \citet[p.~65]{finck_RealParameterBlackBoxOptimization_noiseless}\\
    Different Powers Function & High conditioning and unimodal & \citet[p.~70]{finck_RealParameterBlackBoxOptimization_noiseless}\\
    \midrule
    Rastrigin Function & Multimodal with adequate global structure & \citet[p.~75]{finck_RealParameterBlackBoxOptimization_noiseless}\\
    Weierstrass Function & Multimodal with adequate global structure & \citet[p.~80]{finck_RealParameterBlackBoxOptimization_noiseless}\\
    Schaffers F7 Function & Multimodal with adequate global structure & \citet[p.~85]{finck_RealParameterBlackBoxOptimization_noiseless}\\
    Schaffers F7 Function, moderately ill-conditioned & Multimodal with adequate global structure & \citet[p.~90]{finck_RealParameterBlackBoxOptimization_noiseless}\\
    Composite Griewank-Rosenbrock Function F8F2 & Multimodal with adequate global structure & \citet[p.~95]{finck_RealParameterBlackBoxOptimization_noiseless}\\
    \midrule
    Schwefel Function & Multimodal with weak global structure & \citet[p.~100]{finck_RealParameterBlackBoxOptimization_noiseless}\\
    Katsuura Function & Multimodal with weak global structure & \citet[p.~115]{finck_RealParameterBlackBoxOptimization_noiseless}\\
    Lunacek bi-Rastrigin Function & Multimodal with weak global structure & \citet[p.~120]{finck_RealParameterBlackBoxOptimization_noiseless}\\
    \bottomrule
\end{tabular*}
\end{table*}

\subsection{Black-Box Optimization Tasks}
\label{appendix:bbob-eval}
To rigorously evaluate generalization, we test on six challenging out-of-distribution functions (\Cref{tab:bbob-ood}). These include the two held-out BBOB functions and four additional benchmarks from \citet{jamil_LiteratureSurveyBenchmark_2013}. The selected functions feature different characteristics from the training set, including the Ackley function (known for its narrow global optimum), the Dixon-Price function (which tests optimization in narrow valleys), and the Levy function (characterized by numerous local optima).
\begin{table*}
\caption{Out-of-distribution black-box optimization tasks}
\label{tab:bbob-ood}
\begin{tabular*}{\linewidth}{@{\extracolsep{\fill}} lll }
    \toprule
    Function & Property & Reference\\
    \midrule
    Gallagher's Gaussian 101-me Peaks Function & Multimodal with weak global structure & \citet[p.~105]{finck_RealParameterBlackBoxOptimization_noiseless}\\
    Gallagher's Gaussian 21-hi Peaks Function & Multimodal with weak global structure & \citet[p.~110]{finck_RealParameterBlackBoxOptimization_noiseless}\\
    Ackley Function & Non-Separable, multimodal & \citet[p.~5]{jamil_LiteratureSurveyBenchmark_2013}\\
    Dixon-Price Function & Non-Separable, unimodal & \citet[p.~15]{jamil_LiteratureSurveyBenchmark_2013}\\
    Salomon Function & Non-Separable, multimodal & \citet[p.~27]{jamil_LiteratureSurveyBenchmark_2013}\\
    Levy Function & Non-Separable, multimodal & \citet[p.~40]{jamil_LiteratureSurveyBenchmark_2013}\\
    \bottomrule
\end{tabular*}
\end{table*}

\subsection{Robot Control Tasks}
\label{appendix:robotic}
We evaluate our method on six robotic control tasks implemented in the Brax physics simulator~\citep{brax} and one arm reaching task from~\citet{cully_RobotsThatCan_2015}. These tasks represent significant departures from the training distribution, featuring high-dimensional search spaces (up to 584 parameters), domain-specific behavioral descriptors, and complex fitness landscapes shaped by physics-based constraints.

As detailed in~\Cref{tab:robot}, each environment presents unique challenges. The locomotion tasks (Hopper, Walker, Half Cheetah, and Ant) optimize for forward velocity while using either foot contact patterns or velocity-based behavioral descriptors. The arm reaching task optimizes for precise end-effector positioning while using joint angles as descriptors. These tasks test both the scalability of our method to high-dimensional problems and its ability to leverage meaningful, domain-specific behavioral descriptors.
\begin{table*}
\caption{Robot Control Tasks}
\label{tab:robot}
\begin{tabular*}{\linewidth}{@{\extracolsep{\fill}} lllll }
    \toprule
    Environment & Fitness & Descriptor & Genotype size & Reference\\
    \midrule
    Hopper & Forward velocity & Velocity and feet contact & 243 & \citet{brax}\\
    Walker & Forward velocity & Feet contact & 390 & \citet{brax}\\
    Half Cheetah & Forward velocity & Feet contact & 390 & \citet{brax}\\
    Ant & Forward velocity & Velocity & 584 & \citet{brax}\\
    Ant & Forward velocity & Feet contact & 584 & \citet{brax}\\
    Arm & Standard deviation of joint angles & Final position of end effector & 8 & \citet{cully_RobotsThatCan_2015}\\
    \bottomrule
\end{tabular*}
\end{table*}

\end{document}